\documentclass{article}

\usepackage[preprint]{neurips_2024}
\makeatletter
\renewcommand{\@noticestring}{}
\makeatother

\usepackage[utf8]{inputenc} 
\usepackage[T1]{fontenc}    
\usepackage{hyperref}       
\usepackage{url}            
\usepackage{booktabs}       
\usepackage{multirow}       
\usepackage{amsfonts}       
\usepackage{nicefrac}       
\usepackage{microtype}      
\usepackage{xcolor}         

\usepackage{graphicx}
\usepackage{subfigure}
\usepackage{wrapfig}

\usepackage{amsmath}
\usepackage{amssymb}
\usepackage{mathtools}
\usepackage{enumitem}
\usepackage{bm}

\usepackage[titlenumbered,ruled,linesnumbered,vlined]{algorithm2e}

\usepackage{listings}
\usepackage[capitalize,noabbrev]{cleveref}

\usepackage{xspace}
\newcommand{\centaur}{Centaur\xspace}

\newif\ifshowack
\showacktrue   

\title{Can LLMs Beat Classical Hyperparameter Optimization Algorithms? A Study on \href{https://github.com/karpathy/autoresearch}{\textit{autoresearch}}}

\author{%
  Fabio Ferreira$^{1,2,}$\thanks{Equal supervision. Correspondence to \texttt{fabioferreira <at> mailbox <dot> org}} ,\quad
  Lucca Wobbe$^{3}$,\quad
  Arjun Krishnakumar$^{2}$, \\
  \textbf{Frank Hutter}$^{1,2,4,*}$,\quad
  \textbf{\&}
  \textbf{Arber Zela}$^{1,*}$ \\[0.5em]
  $^{1}$ELLIS Institute T\"ubingen\quad
  $^{2}$University of Freiburg \\
  $^{3}$Karlsruhe Institute of Technology\quad
  $^{4}$Prior Labs
}

\begin{document}
\raggedbottom

\maketitle

\begin{abstract}
The \href{https://github.com/karpathy/autoresearch}{\emph{autoresearch}} repository enables an LLM agent to optimize hyperparameters by editing training code directly. We use it as a testbed to compare classical HPO algorithms against LLM-based methods on tuning the hyperparameters of a small language model under a fixed compute budget.
When defining a fixed search space over autoresearch, classical methods such as CMA-ES and TPE consistently outperform LLM-based agents, where avoiding out-of-memory failures matters more than search diversity. Allowing the LLM to directly edit source code narrows the gap to the classical methods but does not close it, even with frontier models available at the time of writing such as Claude Opus 4.6 and Gemini 3.1 Pro Preview. We observe that LLMs struggle to track optimization state across trials. In contrast, classical methods lack the domain knowledge of LLMs.
To combine the strengths of both, we introduce \centaur, a hybrid that shares CMA-ES's interpretable internal state, including mean vector, step-size, and covariance matrix, with an LLM\footnote{Named after the mythological half-human, half-horse hybrid: our method merges LLM reasoning with classical optimization.}. \centaur achieves the best result in our experiments, and a 0.8B LLM already suffices to outperform all classical and pure LLM methods. Unconstrained code editing requires larger models to be competitive with classical methods. We further analyze search diversity, model scaling from 0.8B to frontier models, and ablate the fraction of LLM-proposed trials in \centaur. All in all, our results suggest that LLMs are most effective as a complement to classical optimizers, not as a replacement.
Code is available at \url{https://github.com/ferreirafabio/autoresearch-automl} \& interactive demo at \url{https://ferreirafabio.github.io/autoresearch-automl}.
\end{abstract}

\section{Introduction}
\label{sec:intro}

\emph{autoresearch}~\citep{karpathy2025autoresearch} demonstrated that an LLM agent can iteratively edit training code to improve a small language model (${\sim}$50M parameters), reaching a val\_bpb of ${\sim}$0.978 in less than 100 trials on a single H100. HPO is a central component of AutoML~\citep{feurer-automlbook19a, bischl-dmkd23a} that can encompass various hyperparameters of a learning pipeline, from model architecture and optimizer settings to data augmentation strategies~\citep{pineda-iclr24a, wagner-icml22a}. This raises two questions: \emph{(i)~how do other classical HPO methods perform on this task?} and \emph{(ii)~can LLM-based HPO methods outperform classical ones?}

\begin{figure}[t]
    \centering
    \includegraphics[width=\textwidth]{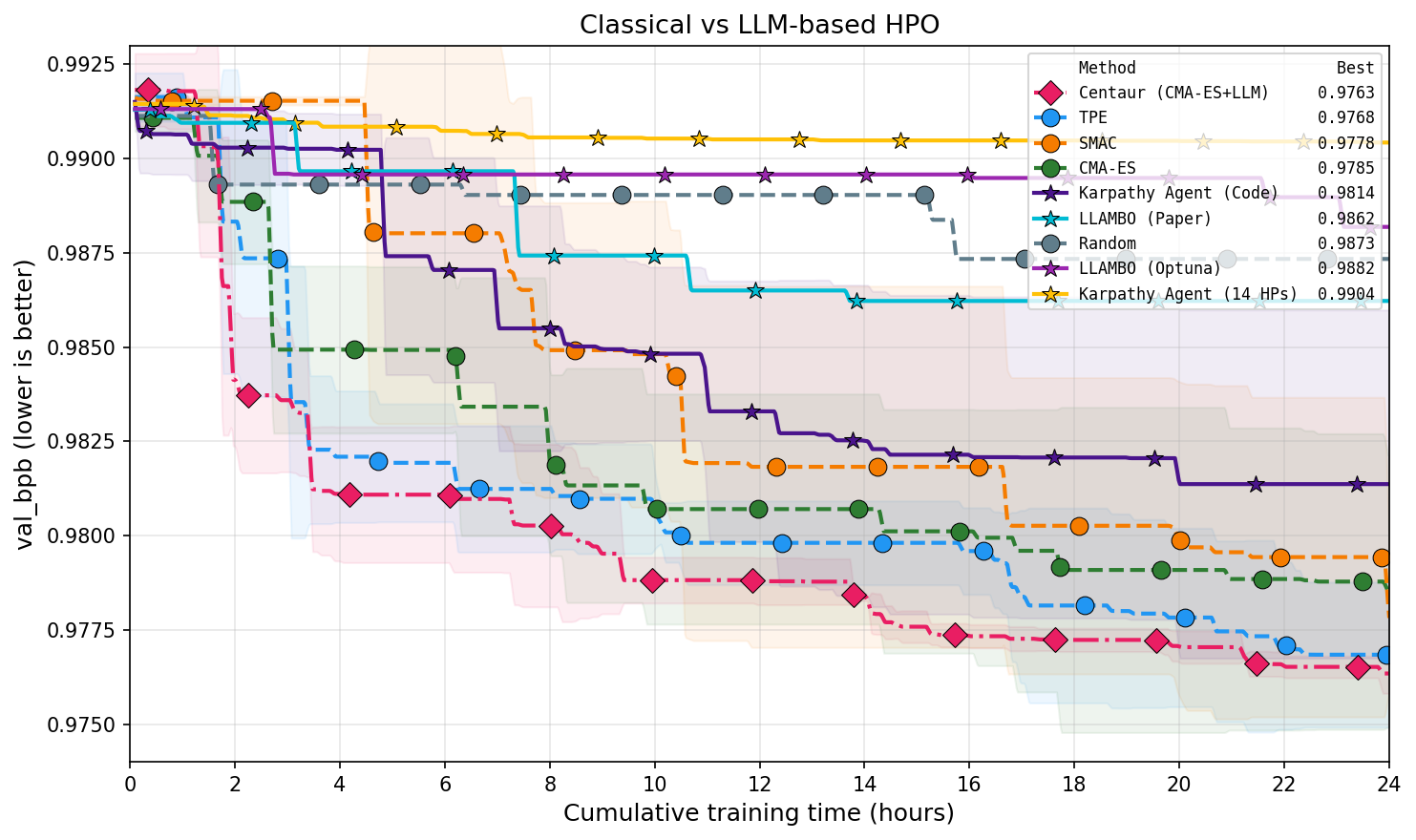}
    \vspace{-0.5em}
    \caption{Best val\_bpb (mean $\pm$ std, 3 seeds) against cumulative training time. All methods receive the same 24-hour GPU budget; LLM inference overhead is excluded. All LLM-based methods use Qwen3.5-27B. Linestyle and markers encode method category: \textbf{dashed with circles} = classical, \textbf{solid with stars} = pure LLM, \textbf{dash-dot with diamonds} = hybrid (\centaur). Marker positions are offset per line for visual clarity. Bands shown for top 5 methods (see \Cref{fig:twopanel} for all).}
    \label{fig:convergence}
\end{figure}

To answer these questions, we benchmark 9 HPO methods, spanning 4 classical, 4 LLM-based, and 1 hybrid, on Karpathy's autoresearch task. All methods operate under the same 24-hour GPU training budget with 3 seeds. To reduce human priors, we automatically extract 14 hyperparameters from the training script; while the ranges require some domain knowledge, the HP selection itself is automated, removing manual search space curation. \Cref{fig:convergence} compares all 9 methods using 27B LLM variants against cumulative training wall-time, showing that classical methods find better configurations than LLM-based agents within the fixed search space. The exception is Karpathy Agent (Code), which directly edits training source code and is competitive with classical methods, even though the latter find a similar performing hyperparameter configuration ${\sim}4{\times}$ faster. All LLM methods use a self-hosted open-weight model (Qwen3.5-27B); experiments with frontier models Gemini 3.1 Pro Preview~\citep{google2026gemini31} and Claude Opus 4.6~\citep{anthropic2026opus} show that pure LLM agents do not outperform classical methods even with frontier models, though scaling benefits hybrid methods more than pure LLM agents (\Cref{fig:frontier_main}).

Beyond this comparison, we propose \centaur, a hybrid method that combines CMA-ES with an LLM by sharing the optimizer's full internal state, including the mean vector~$\bm{\mu}$, step-size~$\sigma$, and covariance matrix~$\mathbf{C}$. The hypothesis is that CMA-ES and LLMs have complementary strengths: CMA-ES learns the optimization landscape but lacks domain knowledge, while the LLM brings transformer training intuitions but, at the small and mid-sized scales we test (0.8B and 27B), struggles to track optimization state across trials reliably, e.g., LLM methods show OOM rates comparable to random search despite observing full trial history. This motivates pairing the LLM with a classical optimizer whose state can be shared explicitly. We chose CMA-ES because its internal state is particularly interpretable for LLM communication (see \Cref{sec:method}).

In summary, we make the following contributions:
\begin{itemize}[nosep,leftmargin=*]
    \item We \textbf{benchmark 9 HPO methods} on autoresearch~\citep{karpathy2025autoresearch}, supporting both fixed-HP and agentic code-editing optimization, under identical 24-hour budgets with 3 seeds.
    \item We show that \textbf{classical HPO outperforms LLM agents} within a fixed search space. An LLM agent that directly edits code is \textbf{more competitive} but still falls short of the best classical methods.
    \item We introduce \textbf{\centaur}, a hybrid that shares CMA-ES's full internal state with the LLM and \textbf{achieves the best result} in our experiments, where a 0.8B LLM already suffices to outperform all classical and pure LLM methods.
    \item We analyze search diversity, OOM rates, and model scaling across all methods. Experiments with frontier models confirm that \textbf{simply scaling the LLM does not outperform} classical methods, though \centaur with Opus 4.6 extends its lead further.
\end{itemize}

\vspace{-0.5em}
    
\section{Related Work}
\label{sec:related}

Classical HPO spans a wide range of approaches~\citep{bischl-dmkd23a, feurer-automlbook19a}, from random search~\citep{bergstra-jmlr12a} and Bayesian optimization with Gaussian process surrogates~\citep{snoek-nips12a} to sequential model-based approaches with random forests such as SMAC~\citep{hutter-lion11a}, tree-structured Parzen estimators such as TPE~\citep{bergstra-nips11a}, and evolution strategies such as CMA-ES~\citep{hansen2016cmaes}. We focus on single-task HPO methods without multi-fidelity or transfer to isolate each optimizer's ability to learn the landscape from scratch. Methods such as Hyperband~\citep{li-iclr17a}, BOHB~\citep{falkner-icml18a}, transfer HPO~\citep{wistuba-iclr21a, pineda-iclr24a}, zero-shot approaches~\citep{ferreira-icml22a}, multi-task transfer~\citep{strangmann-neuripsws24a}, and hyperparameter-robust training~\citep{ferreira-iclr22a} could in principle be applied to this benchmark but are not in the scope of this study.

In line with the increasing interest in open-ended agentic discovery~\citep{zhang2026darwin,Novikov2025AlphaEvolveAC,lange2026shinkaevolve,liu2024llm4ad,wang2026huxleygodel}, recent work explores LLMs as components in HPO pipelines. While in principle, methods such as AlphaEvolve~\citep{Novikov2025AlphaEvolveAC} or ShinkaEvolve~\citep{lange2026shinkaevolve} can be tasked to optimize the objective in autoresearch, we focus only on methods tailored specifically for HPO. LLAMBO~\citep{liu2024llambo} uses an LLM as a surrogate model inside Bayesian optimization, replacing the Gaussian process with LLM-based performance predictions. SLLMBO~\citep{mahammadli2024sllmbo} integrates an LLM with TPE into a joint sampler. \citet{zhang2023llmhpo} prompt LLMs directly for HP suggestions. LLaMA-ES~\citep{kramer2024llamaes} uses LLMs to tune CMA-ES's own hyperparameters. \citet{schwanke2025hollm} partition the search space into subregions via a bandit mechanism and use an LLM to propose candidates within each region. The authors further extend that to multi-objective optimization~\citep{Schwanke2026MultiObjectiveHO}. However, none of these methods share the classical optimizer's full internal state with the LLM, which is the key idea behind \centaur.

Our work differs from prior work in four ways. First, we use autoresearch~\citep{karpathy2025autoresearch} as a real-world benchmark that naturally accommodates both classical HPO within a fixed search space and agentic LLM optimization through direct code editing, enabling a head-to-head comparison under identical conditions. Second, we benchmark classical and LLM methods on this task following best practices for algorithm configuration~\citep{eggensperger-jair19a}. Third, we automatically extract HPs from the training script via Abstract Syntax Tree (AST) parsing to control for human priors in the search space. Fourth, we introduce \centaur, a hybrid that explicitly passes CMA-ES's mean, $\sigma$, and $\mathbf{C}$ to the LLM, enabling optimizer-informed suggestions rather than history-only LLM reasoning. In contrast, LLAMBO uses the LLM as a surrogate where the acquisition function decides, SLLMBO combines LLM and TPE proposals without exposing optimizer internals, and HOLLM~\citep{schwanke2025hollm} uses spatial partitioning to constrain the LLM rather than inform it with optimizer state.

\section{Experimental Setup}
\label{sec:setup}

We describe the benchmark task and hardware, then the evaluation protocol and failure handling, followed by the LLM infrastructure, and finally the automated search space extraction.

We evaluate all methods on nanochat~\citep{karpathy2025nanochat}, a small decoder-only transformer~\citep{radford-openaiblog19a} trained on FineWeb~\citep{penedo-neurips24a}, optimizing validation bits-per-byte (val\_bpb). Each trial ran for five minutes on a single NVIDIA H200 GPU (141\,GB HBM3e).

We ran each method for 24 hours with three seeds and report results against cumulative training wall-time. Methods with high OOM rates accumulated more trials due to fast failures (${<}30$s per OOM); trial-number plots in the appendix are capped at 300 trials as no meaningful improvement occurs beyond that point. We report failed trials (OOM) as val\_bpb${}=100.0$, a finite penalty orders of magnitude worse than any valid result while remaining compatible with all surrogate models, so that optimizers learn to avoid infeasible regions.

Our main experiments use Qwen3.5~\citep{qwen35blog} (0.8B and 27B) as the \emph{LLM optimizer}, self-hosted via vLLM~\citep{kwon2023vllm} on the same GPU that trains the \emph{optimizee} (the ${\sim}$50M-parameter language model). We additionally run frontier model experiments with Gemini 2.5 Flash~\citep{comanici2025gemini}, 3.1 Flash-Lite, and 3.1 Pro Preview~\citep{google2026gemini31} via the Gemini API. To ensure equal resource allocation, we capped training VRAM to 80\,GB for all methods, comparable to the H100 used by Karpathy~\citep{karpathy2025autoresearch}, and reserved the remaining memory for the vLLM server when using self-hosted models. We disabled Qwen3.5's thinking mode and sampled with temperature 0.7, limiting outputs to 2048 tokens for fixed-HP methods and 16384 tokens for code editing. Each LLM-based method receives a prompt containing the optimization goal (minimize val\_bpb), the model class (GPT-2 scale transformer with Muon+AdamW optimizer), the dataset (FineWeb), the hardware (H200 GPU with VRAM budget and OOM warning), the search space with bounds, and the trial history including both successful and failed configurations. Full prompt templates for all five LLM-based methods are reproduced verbatim in \Cref{sec:prompts}.

To extract hyperparameters from \texttt{train.py}, we used Abstract Syntax Tree (AST) parsing: we walk through parse tree and collect all top-level assignments whose variable name is in \texttt{ALL\_CAPS} format and have a right-hand as a literal value (integer, float, or string). This yielded 14 HPs (13 continuous/integer and one categorical), listed in \Cref{tab:searchspace} with their types, ranges, and defaults from Karpathy's starting configuration. While the ranges require some domain knowledge, the hyperparameter selection itself is fully automated in this way, removing manual search space curation. \Cref{tab:methods} summarizes the nine methods we evaluated across four classical, four LLM-based, and one hybrid approach.

\begin{table}[t]
\centering
\caption{Search space: 14 HPs auto-extracted via AST parsing. Ranges are set manually. Defaults are Karpathy's starting config. \texttt{WINDOW\_PATTERN} controls the per-layer attention window: S = short (local) attention, L = long (full) attention.}
\label{tab:searchspace}
\small
\begin{tabular}{@{}lllcl@{}}
\toprule
HP & Type & Range & Log & Default \\
\midrule
\texttt{DEPTH} & int & 4--24 & & 8 \\
\texttt{ASPECT\_RATIO} & int & 32--128 & & 64 \\
\texttt{HEAD\_DIM} & int & 64--256 & \checkmark & 128 \\
\texttt{DEVICE\_BATCH\_SIZE} & int & 32--256 & \checkmark & 128 \\
\texttt{TOTAL\_BATCH\_SIZE} & int & 65\,536--2\,097\,152 & \checkmark & 524\,288 \\
\texttt{EMBEDDING\_LR} & float & 0.01--2.0 & \checkmark & 0.6 \\
\texttt{UNEMBEDDING\_LR} & float & 0.0005--0.05 & \checkmark & 0.004 \\
\texttt{MATRIX\_LR} & float & 0.005--0.2 & \checkmark & 0.04 \\
\texttt{SCALAR\_LR} & float & 0.05--2.0 & \checkmark & 0.5 \\
\texttt{WEIGHT\_DECAY} & float & 0.0--0.5 & & 0.2 \\
\texttt{WARMUP\_RATIO} & float & 0.0--0.3 & & 0.0 \\
\texttt{WARMDOWN\_RATIO} & float & 0.1--0.8 & & 0.5 \\
\texttt{FINAL\_LR\_FRAC} & float & 0.0--0.2 & & 0.0 \\
\texttt{WINDOW\_PATTERN} & cat. & \{SSSL, SSLL, SLSL, LLLL, SSSS, LSSL\} & & SSSL \\
\bottomrule
\end{tabular}
\end{table}

\begin{table}[b]
\centering
\caption{Methods evaluated. ``Fixed'' indicates the shared 14-HP search space; ``Unconstrained'' indicates direct source code editing.}
\label{tab:methods}
\footnotesize
\begin{tabular}{@{}llp{6.5cm}@{}}
\toprule
Method & Search Space & Description \\
\midrule
TPE & Fixed & Tree-structured Parzen Estimator via Optuna~\citep{bergstra-nips11a, akiba-kdd19a} \\
CMA-ES & Fixed & Covariance Matrix Adaptation via Optuna's CMA sampler~\citep{hansen2016cmaes} \\
SMAC & Fixed & Random forest surrogate~\citep{hutter-lion11a} \\
Random & Fixed & Uniform random sampling \\
\midrule
LLAMBO (Optuna) & Fixed & OptunaHub~\citep{ozaki2025optunahub} port: binary surrogate labels, random categorical sampling \\
LLAMBO (Paper) & Fixed & Reimplementation faithful to \citet{liu2024llambo}: continuous labels, all HPs visible \\
Karpathy Agent (14 HPs) & Fixed & LLM sees trial history, suggests configs within fixed space \\
Karpathy Agent (Code) & Unconstrained & LLM directly edits \texttt{train.py} source code~\citep{karpathy2025autoresearch} \\
\midrule
\centaur & Fixed & CMA-ES shares internal state with LLM (\Cref{sec:method}) \\
\bottomrule
\end{tabular}
\end{table}

\section{\centaur: CMA-ES Guided LLM Optimization}
\label{sec:method}

\setlength{\intextsep}{0pt}%
\begin{wrapfigure}{r}{0.51\textwidth}
\vspace{-.1em}
\begin{minipage}{0.51\textwidth}
\begin{algorithm}[H]
\caption{\centaur}
\label{alg:centaur}
\small
\DontPrintSemicolon
\LinesNotNumbered
\KwIn{Search space $\mathcal{S}$, budget $T$, LLM ratio $r{=}0.3$}
Initialize CMA-ES, history $\mathcal{H} \leftarrow \emptyset$\;
\For{$t = 1, \dots, T$}{
    \eIf{LLM turn (with probability $r$)}{
        Extract $\bm{\mu}, \sigma, \mathbf{C}$ from CMA-ES\;
        $\mathbf{x} \leftarrow$ LLM($\bm{\mu}, \sigma, \mathbf{C}$, $\mathcal{H}$, $\mathcal{S}$)\;
    }{
        $\mathbf{x} \leftarrow$ CMA-ES.Propose()\;
    }
    $y \leftarrow$ Evaluate($\mathbf{x}$)\;
    CMA-ES.Update($\mathbf{x}, y$)\;
    $\mathcal{H} \leftarrow \mathcal{H} \cup \{(\mathbf{x}, y)\}$\;
}
\end{algorithm}
\vspace{-0.2em}
\end{minipage}
\end{wrapfigure}
\centaur shares CMA-ES's full internal state with the LLM on a fraction of trials. On every trial, CMA-ES proposes a candidate configuration from its multivariate Gaussian, parameterized by mean vector $\bm{\mu}$, step-size $\sigma$, and covariance matrix $\mathbf{C}$. On 30\% of trials, the LLM receives CMA-ES's proposal along with $\bm{\mu}$, $\sigma$, $\mathbf{C}$, the top-5 configurations, and the last 20 trials. 
The LLM may override the proposal, and in practice does so in nearly all cases: 100\% of the time with the 27B model and 95\% with the 0.8B model.
Crucially, CMA-ES updates its internal state from all trial results, including those where the LLM overrode its proposal, so the optimizer continuously learns from the full trajectory.

We chose CMA-ES because its internal state is particularly interpretable for LLM communication: the mean is a concrete configuration, $\sigma$ is a single scalar, and $\mathbf{C}$ is a labeled matrix. In contrast, TPE maintains density estimators that are difficult to summarize in natural language, and GP-BO maintains a high-dimensional posterior.

\section{Results}
\label{sec:results}

We compare classical and LLM-based HPO methods across three axes: fixed search space performance, unconstrained code editing, and hybrid optimization. All methods started from the same baseline configuration (Karpathy's default, val\_bpb${\approx}0.991$). We exclude LLM inference overhead from wall-time to isolate optimization quality from inference cost.

\subsection{Classical methods outperform LLMs in fixed search spaces}
\label{sec:results_fixed}

\Cref{fig:convergence} shows convergence against cumulative training wall-time for all 27B methods. Within the fixed search space, classical HPO methods consistently outperformed pure LLM-based approaches. The top methods by mean best val\_bpb are \centaur (0.9763), TPE (0.9768), SMAC (0.9778), CMA-ES (0.9785), and Karpathy Agent (Code) (0.9814). The gap to the best fixed-space LLM method (LLAMBO (Paper)) is significant, and several pure LLM methods performed worse than random search, indicating that restricting LLMs to a fixed HP search space does not leverage their strengths.

Moreover, OOM avoidance matters more than search diversity: \Cref{tab:diversity} shows that the top methods all have OOM rates at or below 16\%, while the bottom four exceed 36\%. Karpathy Agent (14 HPs) has the lowest diversity by all metrics, converging to a narrow region early. LLAMBO (Paper) is one of the most diverse methods yet underperforms classical methods due to its 48\% OOM rate. These OOM rates also illustrate the state-tracking limitation of small and mid-sized LLMs: LLAMBO (Paper) and LLAMBO (Optuna) observe full trial history yet produce OOM rates (48\% and 61\%) comparable to random search (56\%), suggesting they fail to learn which regions of the search space cause memory failures. In contrast, CMA-ES and TPE maintain explicit optimization state and keep OOM rates at 16\% and 11\%, showing that covariance adaptation and density estimation learn which regions are safe. However, CMA-ES also has the highest variance among top methods (std 0.0036 vs 0.0019 for TPE), with its best seed achieving the single best result in the benchmark (0.9741) while its worst seed is mediocre (0.9829).

We report both LLAMBO variants because the OptunaHub implementation differs from the original paper in how it handles surrogate labels, categorical HPs, and failed trials, which substantially affects OOM rates and performance (see \Cref{sec:llambo_comparison} for details).

\begin{table}[t]
\centering
\caption{Search diversity analysis (3 seeds, alphabetical). ``KA'' abbreviates ``Karpathy Agent''; LLM-based methods are labeled with their LLM optimizer in brackets as in \Cref{fig:convergence,fig:frontier_main}. Best within each category is bold and colored: \textcolor[HTML]{0072B2}{\textbf{hybrid}}, \textcolor[HTML]{E69F00}{\textbf{classical}}, \textcolor[HTML]{009E73}{\textbf{pure LLM}}. \textbf{Trials}: mean${\pm}$std total per seed. \textbf{Best}: mean${\pm}$std best val\_bpb. \textbf{Spread}: mean per-HP std. \textbf{Step}: mean L2 between consecutive trials. Pairwise distance and unique-cell coverage metrics are defined in \Cref{sec:diversity_metrics} and reported for 0.8B variants in \Cref{tab:diversity_08b}.}
\label{tab:diversity}
\footnotesize
\begin{tabular}{@{}llclcc@{}}
\toprule
Method & Trials & Best val\_bpb & OOM\% & Spread & Step \\
\midrule
\centaur\ [Qwen 0.8B] & 333${\pm}$4 & 0.9766${\pm}$0.0008 & 13\% & 0.131 & 0.368 \\
\centaur\ [Qwen 27B] & 334${\pm}$5 & 0.9763${\pm}$0.0005 & 15\% & 0.115 & 0.341 \\
\centaur\ [Gemini 3.1 Pro] & 344${\pm}$10 & 0.9767${\pm}$0.0013 & 20\% & 0.135 & 0.496 \\
\textcolor[HTML]{0072B2}{\textbf{\centaur\ [Opus 4.6]}} & \textcolor[HTML]{0072B2}{\textbf{340${\pm}$13}} & \textcolor[HTML]{0072B2}{\textbf{0.9739${\pm}$0.0012}} & \textcolor[HTML]{0072B2}{\textbf{17\%}} & \textcolor[HTML]{0072B2}{\textbf{0.138}} & \textcolor[HTML]{0072B2}{\textbf{0.499}} \\
CMA-ES & 336${\pm}$13 & 0.9785${\pm}$0.0036 & 16\% & 0.158 & 0.580 \\
KA (14 HPs) [Qwen 27B] & 289${\pm}$2 & 0.9904${\pm}$0.0002 & 1\% & 0.035 & 0.057 \\
KA (Code) [Qwen 0.8B] & 369${\pm}$34 & 0.9910${\pm}$0.0001 & 19\% & -- & -- \\
KA (Code) [Qwen 27B] & 324${\pm}$7 & 0.9814${\pm}$0.0046 & 12\% & -- & -- \\
KA (Code) [Gemini 3.1 Pro] & 308${\pm}$11 & 0.9826${\pm}$0.0004 & 3\% & -- & -- \\
\textcolor[HTML]{009E73}{\textbf{KA (Code) [Opus 4.6]}} & \textcolor[HTML]{009E73}{\textbf{302${\pm}$9}} & \textcolor[HTML]{009E73}{\textbf{0.9770${\pm}$0.0027}} & \textcolor[HTML]{009E73}{\textbf{5\%}} & \textcolor[HTML]{009E73}{\textbf{--}} & \textcolor[HTML]{009E73}{\textbf{--}} \\
LLAMBO (Optuna) [Qwen 27B] & 629${\pm}$64 & 0.9882${\pm}$0.0012 & 61\% & 0.251 & 1.130 \\
LLAMBO (Paper) [Qwen 27B] & 496${\pm}$6 & 0.9862${\pm}$0.0041 & 48\% & 0.252 & 1.218 \\
Random & 568${\pm}$21 & 0.9873${\pm}$0.0021 & 56\% & 0.273 & 1.382 \\
SMAC & 431${\pm}$2 & 0.9778${\pm}$0.0020 & 36\% & 0.238 & 0.369 \\
\textcolor[HTML]{E69F00}{\textbf{TPE}} & \textcolor[HTML]{E69F00}{\textbf{317${\pm}$12}} & \textcolor[HTML]{E69F00}{\textbf{0.9768${\pm}$0.0019}} & \textcolor[HTML]{E69F00}{\textbf{11\%}} & \textcolor[HTML]{E69F00}{\textbf{0.196}} & \textcolor[HTML]{E69F00}{\textbf{0.413}} \\
\bottomrule
\end{tabular}
\end{table}

\begin{figure}[t]
    \centering
    \includegraphics[width=\textwidth]{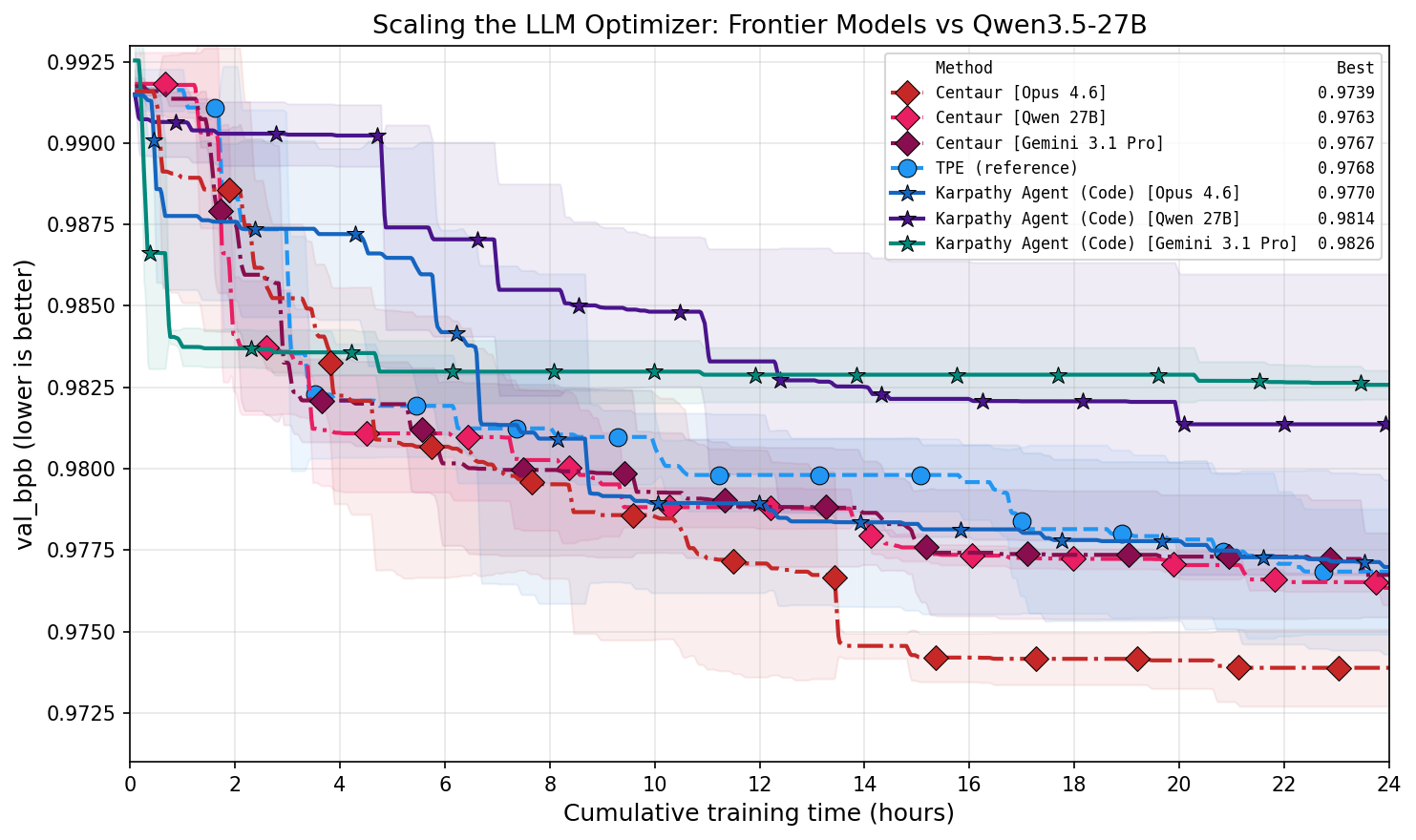}
    \caption{Scaling the LLM optimizer: frontier models vs Qwen3.5-27B for Karpathy Agent (Code) and \centaur (mean $\pm$ std across 3 seeds). Linestyle encodes method category as in \Cref{fig:convergence}. \centaur [Opus 4.6] achieves the best result (0.9739), extending the hybrid method's lead.}
    \label{fig:frontier_main}
\end{figure}

\subsection{Unconstrained code editing is viable but requires model scale}
\label{sec:results_code}

Karpathy Agent (Code), which directly edits training source code rather than operating in the fixed search space, is the only pure LLM method competitive with classical approaches. Given the simplicity of the setup and the use of a self-hosted open-weight model (Qwen3.5-27B), the gap to classical methods is smaller than one might expect.

Scaling the LLM does not consistently close the gap. Comparisons with Gemini 2.5 Flash~\citep{comanici2025gemini} and Gemini 3.1 Flash-Lite~\citep{google2026gemini31} as the LLM optimizer (\Cref{sec:gemini}) show no improvement over Qwen3.5-27B. Gemini 3.1 Flash-Lite fails to generate valid code in 87--94\% of trials for unconstrained editing, but performs comparably to Qwen3.5-0.8B in the fixed search space, confirming that code editing is where model capability matters most. Gemini 3.1 Pro Preview (\Cref{fig:frontier_main}) is competitive with Qwen3.5-27B for both unconstrained code editing (0.9826 vs 0.9814) and \centaur (0.9767 vs 0.9763), with lower OOM rates for code editing (3\% vs 12\%), but does not close the gap to the best classical methods. Claude Opus 4.6~\citep{anthropic2026opus} narrows the gap for code editing to 0.9770${\pm}$0.0027 (vs 0.9814 with Qwen3.5-27B), making it competitive with CMA-ES (0.9785), but still falls short of TPE (0.9768${\pm}$0.0019). Frontier model capability also shows up in the OOM rate (\Cref{tab:diversity}): Karpathy Agent (Code)'s failure rate drops sharply with LLM capability (19\% for 0.8B, 12\% for 27B, 3\% for Gemini 3.1 Pro, 5\% for Opus 4.6), whereas \centaur's stays in a narrow range (13\%--20\%) across model choices. This indicates that for pure code editing, LLM capability directly controls memory-awareness of generated code, while for the hybrid \centaur the CMA-ES side dominates OOM behavior. For pure LLM code editing, scaling the model helps but does not surpass the best classical methods.

\begin{figure}[t]
    \centering
    \includegraphics[width=\textwidth]{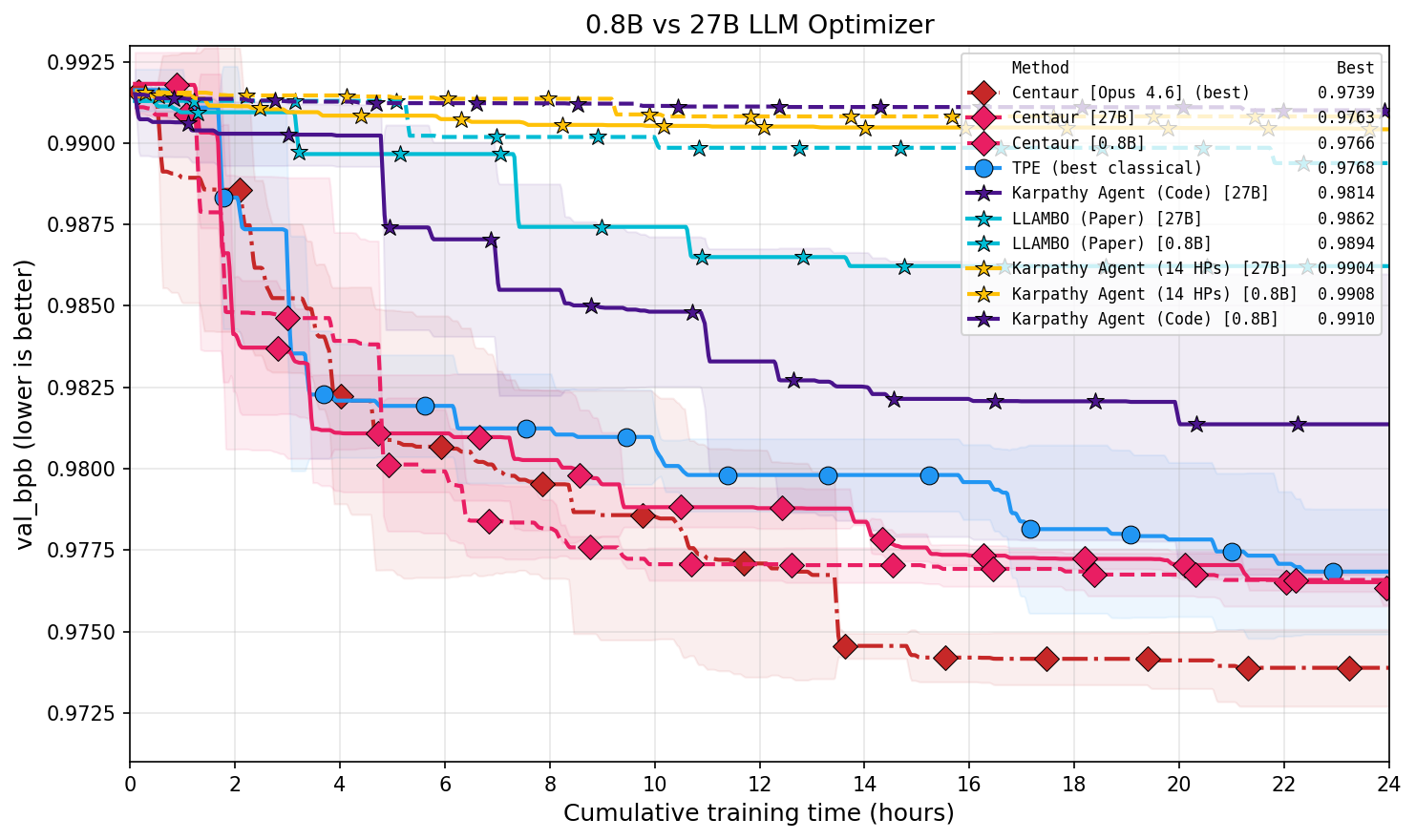}
    \caption{0.8B vs 27B LLM optimizer comparison (wall-time). Solid: 27B, dashed: 0.8B. TPE and Random shown as classical references. The 0.8B model appears insufficient for unconstrained code editing but sufficient for hybrid optimization.}
    \label{fig:modelsize}
\end{figure}

Scaling the LLM from 0.8B to 27B is essential for unconstrained code editing but provides no advantage for fixed-HP optimization, as can be seen in \Cref{fig:modelsize}: 0.8B is insufficient for unconstrained code editing (Karpathy Agent Code: 0.9910 vs 0.9814 with 27B), while fixed-HP methods saw no benefit from scaling (Karpathy Agent 14 HPs: 0.9904 vs 0.9908). All LLM methods in \Cref{fig:convergence,fig:modelsize} used open-weight Qwen3.5.

\subsection{Hybrid optimization: best of both worlds}
\label{sec:results_hybrid}

\centaur outperformed all methods including CMA-ES alone by using the LLM on only 30\% of trials. As described in \Cref{sec:method}, the LLM receives CMA-ES's full internal state ($\bm{\mu}$, $\sigma$, $\mathbf{C}$), the top-5 configurations, and the last 20 trials, and almost always overrides CMA-ES's proposal (100\% for 27B, 95\% for 0.8B). Despite constituting only 30\% of trials, LLM trials contributed 25\% of incumbent improvements (\Cref{sec:centaur_analysis}), confirming that the LLM provides complementary value beyond what CMA-ES finds alone. Beyond improving the mean, \centaur substantially reduces CMA-ES's cross-seed variance: std drops from 0.0036 for CMA-ES alone to 0.0005 for \centaur, suggesting that the LLM stabilizes the optimizer by injecting domain knowledge that prevents unfavorable seeds from drifting. Within the open-weight models tested, \centaur [0.8B] outperformed \centaur [27B], and Gemini 3.1 Pro Preview performed on par with both (0.9766, 0.9763, 0.9767), suggesting a capability plateau where a cheap LLM suffices when paired with a strong classical optimizer. However, Claude Opus 4.6~\citep{anthropic2026opus} breaks through this plateau: \centaur [Opus 4.6] achieves 0.9739${\pm}$0.0012, improving over the open-weight variants by 0.0024 and establishing a new benchmark leader (\Cref{fig:frontier_main}). This improvement is not explained by better OOM avoidance (17\% vs 15\% for Qwen 27B) or more trials, but by higher-quality suggestions during the LLM's 30\% of turns. This indicates that the plateau is not fundamental: a sufficiently capable frontier model provides additional gains even in the hybrid setting, though the practical benefit of open-weight models remains that they achieve competitive performance at a fraction of the inference cost. We ablate the LLM ratio $r$ (see Appendix \Cref{tab:ratio_ablation,fig:ratio_ablation}). For the 0.8B model, $r{=}0.1$ achieves the best result (0.9744), slightly outperforming the default $r{=}0.3$ (0.9766), while $r{=}0.8$ degrades to 0.9849. For the 27B model, $r{=}0.5$ performs best (0.9746), but $r{=}0.8$ collapses to 0.9902, which is worse than CMA-ES alone. This confirms that CMA-ES should retain majority control of the optimization trajectory: the LLM is most effective as an occasional informed perturbation, not as the primary search driver.

\section{Conclusion}
\label{sec:conclusion}

We benchmarked classical, LLM-based, and hybrid HPO methods on small-scale language model training under identical budgets. Within a fixed search space, classical methods consistently outperformed LLM-based agents, with OOM avoidance mattering more than search diversity. Restricting LLMs to a fixed hyperparameter search space does not leverage their strengths; allowing the LLM to directly edit source code narrows the gap but does not close it, even with frontier models such as Claude Opus 4.6 and Gemini 3.1 Pro Preview. \centaur achieved the best result in our experiments by using the LLM on only 30\% of trials, where a 0.8B LLM already suffices for hybrid optimization while unconstrained code editing requires larger models. \centaur with Opus 4.6 extends this lead further, though the improvement comes from higher-quality suggestions rather than better OOM avoidance. All in all, our results suggest that LLMs are most effective as a complement to classical optimizers, not as a replacement.

Our study evaluated a single task with open-weight models (Qwen3.5 0.8B and 27B) and two frontier models (Gemini 3.1 Pro Preview and Claude Opus 4.6). Scaling from 0.8B to 27B to Gemini Pro yields no improvement for hybrid optimization, but \centaur with Opus 4.6 breaks through this plateau. More benchmarking and experiments are needed to determine the generality of these findings. All LLM methods in this benchmark operate without tools or external knowledge retrieval. Equipping LLM optimizers with agentic capabilities such as paper search, documentation lookup, or code analysis tools is a promising direction. Additionally, exploring other classical optimizers as the hybrid base and pairing CMA-ES with a code-editing LLM agent could allow the search space to co-evolve with the optimization trajectory.

\ifshowack
\begin{ack}
This research was funded by the Deutsche Forschungsgemeinschaft (DFG, German Research Foundation) under grant number 539134284, through EFRE (FEIH\_2698644) and the state of Baden-W\"urttemberg. We acknowledge funding by the European Union (via ERC Consolidator Grant DeepLearning 2.0, grant no.~101045765). Views and opinions expressed are however those of the author(s) only and do not necessarily reflect those of the European Union or the European Research Council. Neither the European Union nor the granting authority can be held responsible for them. Frank Hutter acknowledges the financial support of the Hector Foundation.
\begin{center}
\includegraphics[height=0.6cm]{figures/BaWue_Logo_Standard_rgb_pos.png}\hspace{2em}%
\includegraphics[height=0.6cm]{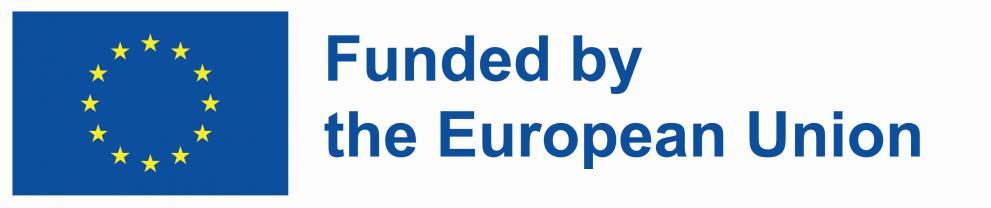}
\end{center}
\end{ack}
\fi

\bibliographystyle{plainnat}
\bibliography{strings,local,lib,shortproc}

\clearpage
\newpage
\appendix
\section{Appendix}
\label{sec:appendix}

\subsection{Convergence by Trial Number}

\begin{figure}[ht!]\vspace{0.5em}
    \centering
    \includegraphics[width=0.9\textwidth]{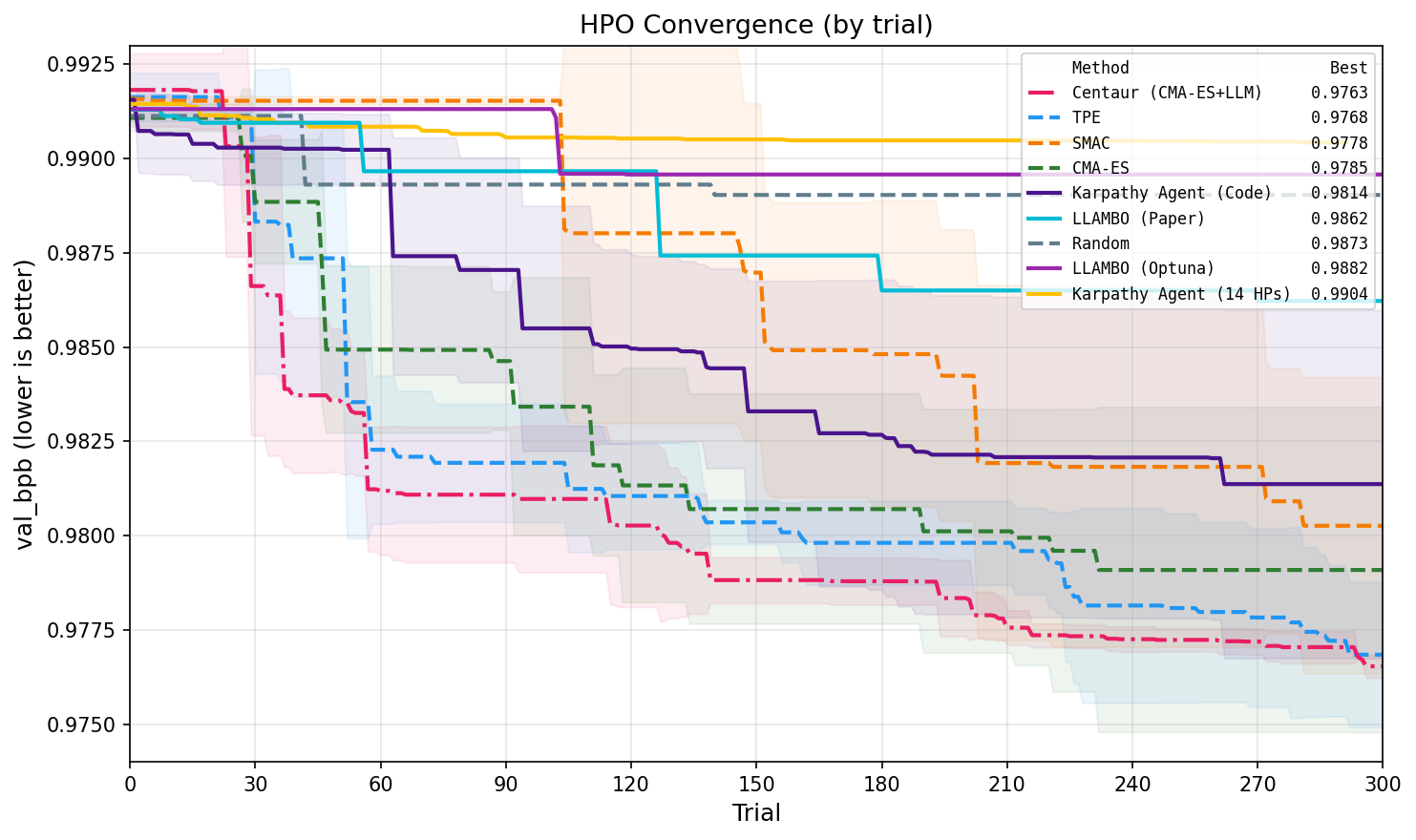}
    \caption{Convergence by trial number (mean $\pm$ std across 3 seeds). Same methods as \Cref{fig:convergence}. Trial-number view shows sample efficiency rather than wall-clock cost.}
    \label{fig:convergence_trial}
\end{figure}

\begin{figure}[ht!]\vspace{0.5em}
    \centering
    \includegraphics[width=0.9\textwidth]{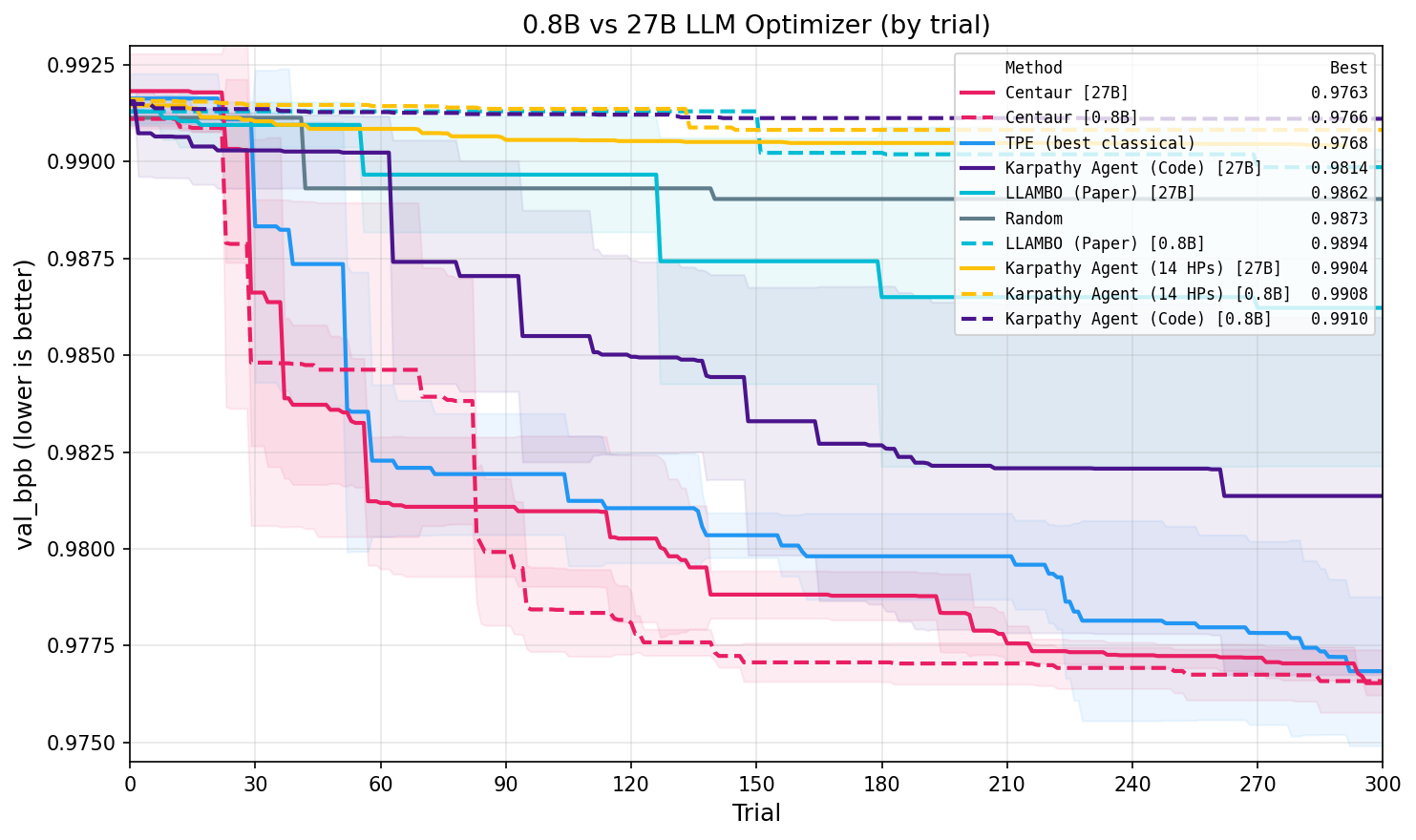}
    \caption{0.8B vs 27B by trial number (mean $\pm$ std across 3 seeds). Same methods as \Cref{fig:modelsize}.}
    \label{fig:modelsize_trial}
\end{figure}

The main text reports convergence against cumulative training time, which is our primary comparison. \Cref{fig:convergence_trial,fig:modelsize_trial} show convergence by trial number instead, measuring sample efficiency. These views can differ substantially because LLM-based methods spend additional time on inference between trials, compressing their wall-time curves even when they are competitive per trial.

\subsection{Per-Seed Convergence}

\begin{figure}[ht!]\vspace{0.5em}
    \centering
    \includegraphics[width=\textwidth]{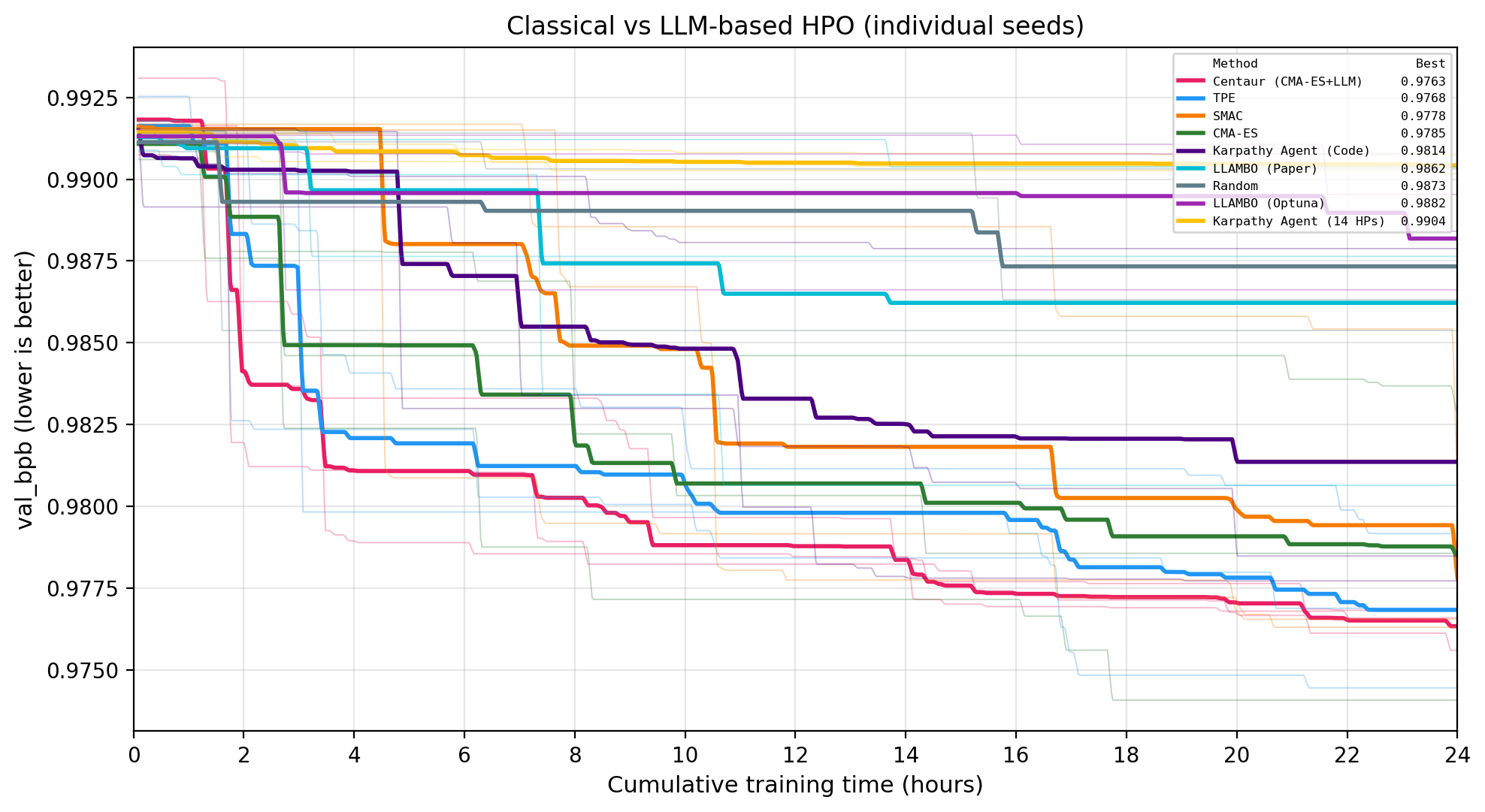}
    \caption{Per-seed convergence for all 27B methods (wall-time). Thin lines: individual seeds. Thick lines: mean across 3 seeds.}
    \label{fig:27b_seeds}
\end{figure}

\begin{figure}[ht!]\vspace{0.5em}
    \centering
    \includegraphics[width=\textwidth]{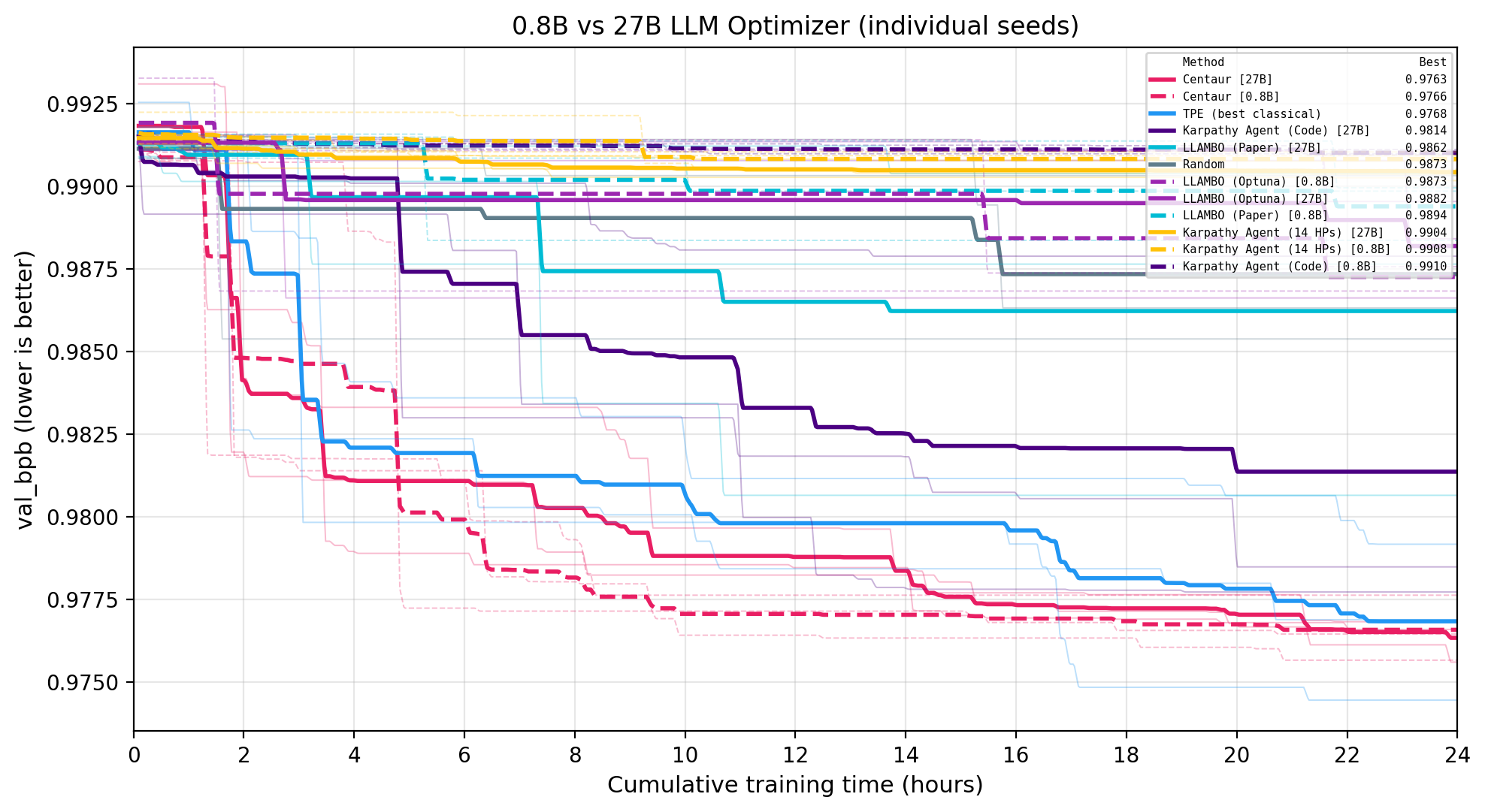}
    \caption{Per-seed convergence for 0.8B vs 27B comparison (wall-time).}
    \label{fig:all_seeds}
\end{figure}

\begin{figure}[ht!]\vspace{0.5em}
    \centering
    \includegraphics[width=\textwidth]{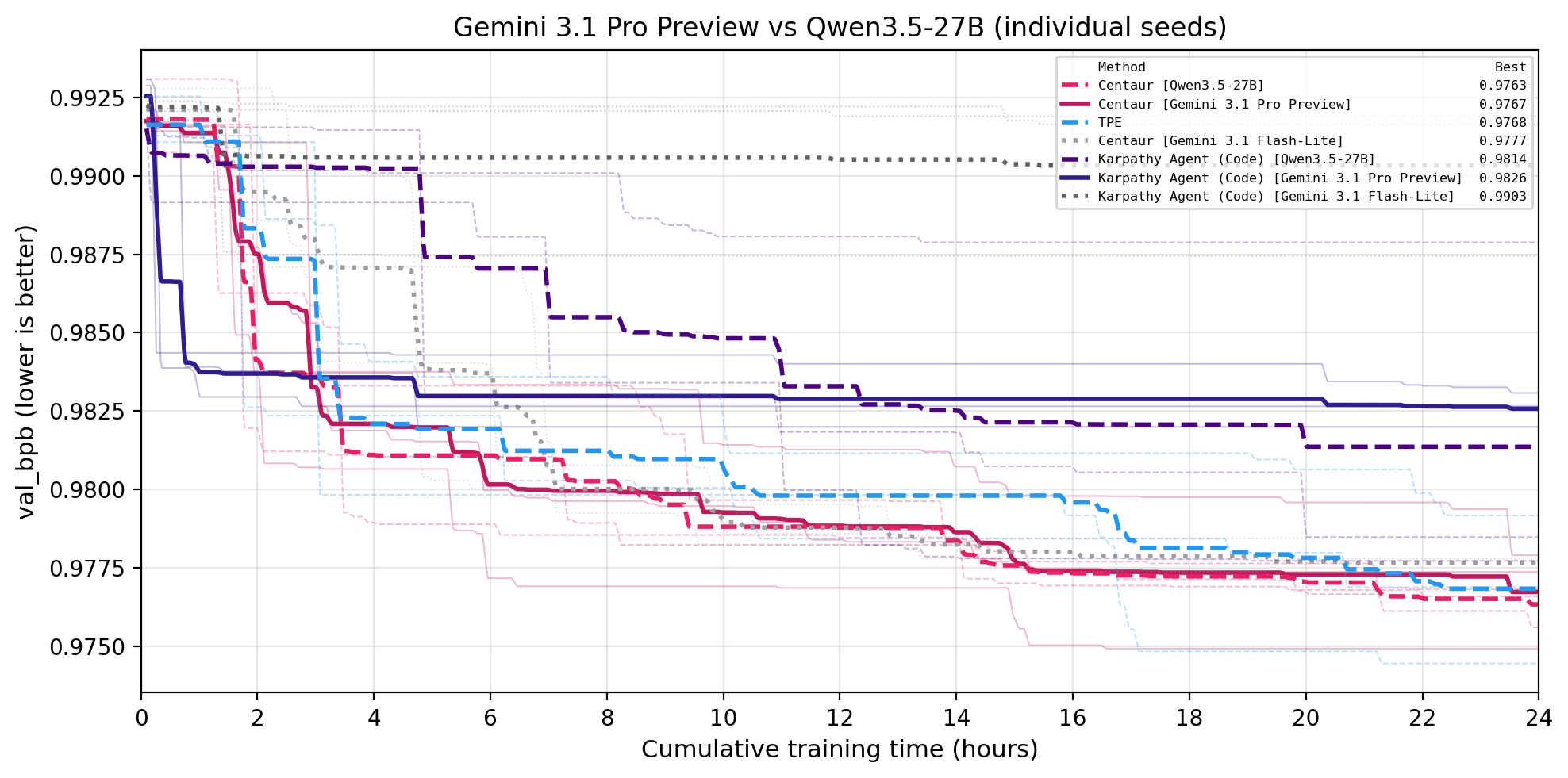}
    \caption{Per-seed convergence for Gemini 3.1 Pro Preview vs Qwen3.5-27B (wall-time).}
    \label{fig:frontier_seeds}
\end{figure}

\Cref{fig:27b_seeds,fig:all_seeds,fig:frontier_seeds} show the same convergence plots as the main text but with individual seed trajectories (thin, transparent) overlaid on the mean (thick, solid) instead of $\pm$std bands. This reveals cross-seed variance and whether the mean is driven by one outlier seed or consistent behavior.

\subsection{Frontier Model Comparison}
\label{sec:gemini}

To test whether a stronger LLM optimizer changes the balance between classical and LLM-based methods, we ran Karpathy Agent (Code) and \centaur with three Gemini variants: Gemini 2.5 Flash~\citep{comanici2025gemini}, Gemini 3.1 Flash-Lite~\citep{google2026gemini31}, and Gemini 3.1 Pro Preview~\citep{google2026gemini31}. Gemini 2.5 Flash and 3.1 Flash-Lite runs were stopped early after 16--18 hours of cumulative training time as no meaningful improvement was observed beyond that point. Gemini 3.1 Pro Preview was run for the full 24-hour budget with 3 seeds.

\begin{figure}[ht!]\vspace{0.5em}
    \centering
    \includegraphics[width=\textwidth]{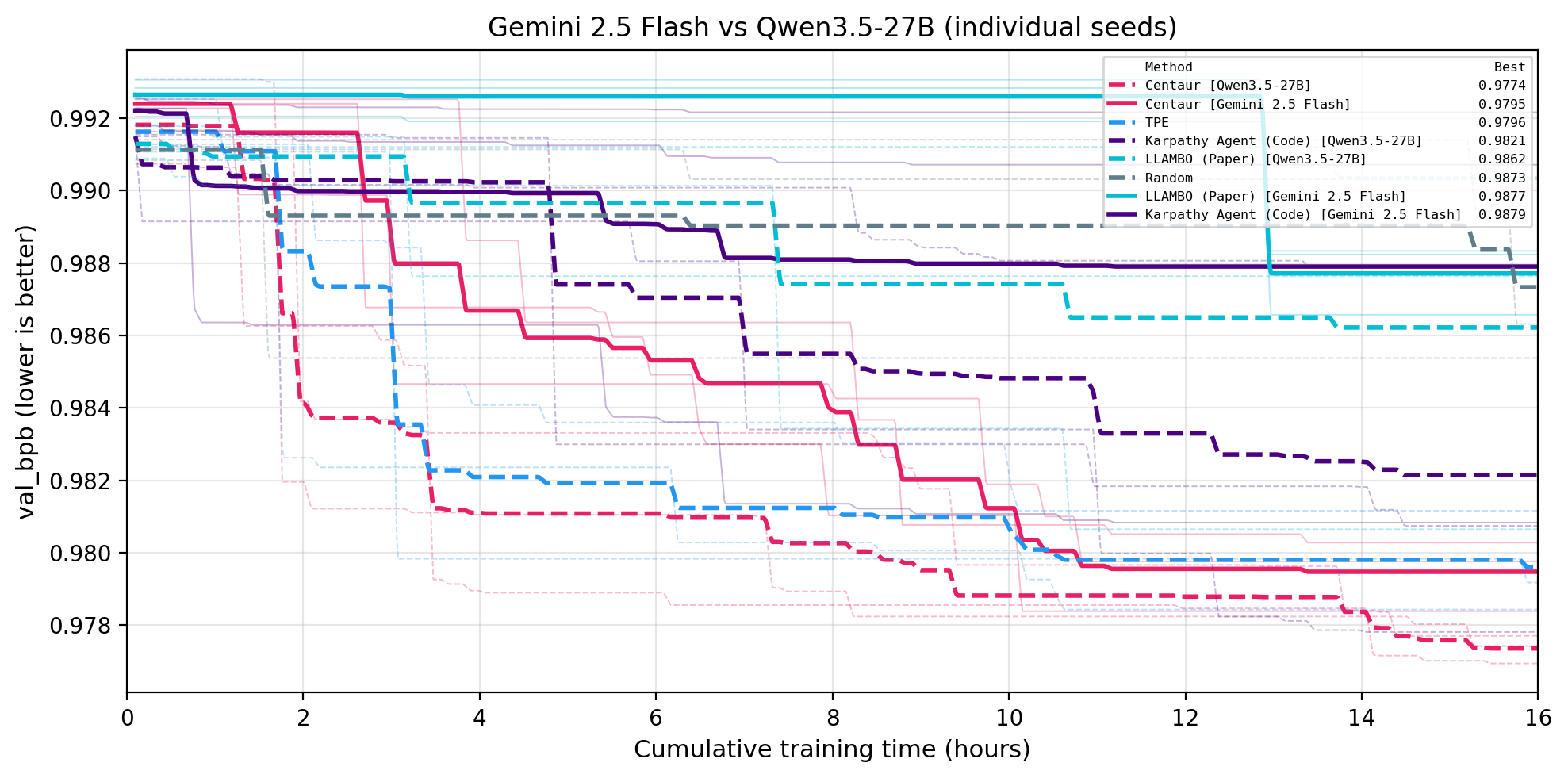}
    \caption{Gemini 2.5 Flash vs Qwen3.5-27B as LLM optimizer (cumulative training time). Solid: Gemini, dashed: Qwen3.5-27B. Same color per method.}
    \label{fig:gemini}
\end{figure}

\begin{figure}[ht!]\vspace{0.5em}
    \centering
    \includegraphics[width=\textwidth]{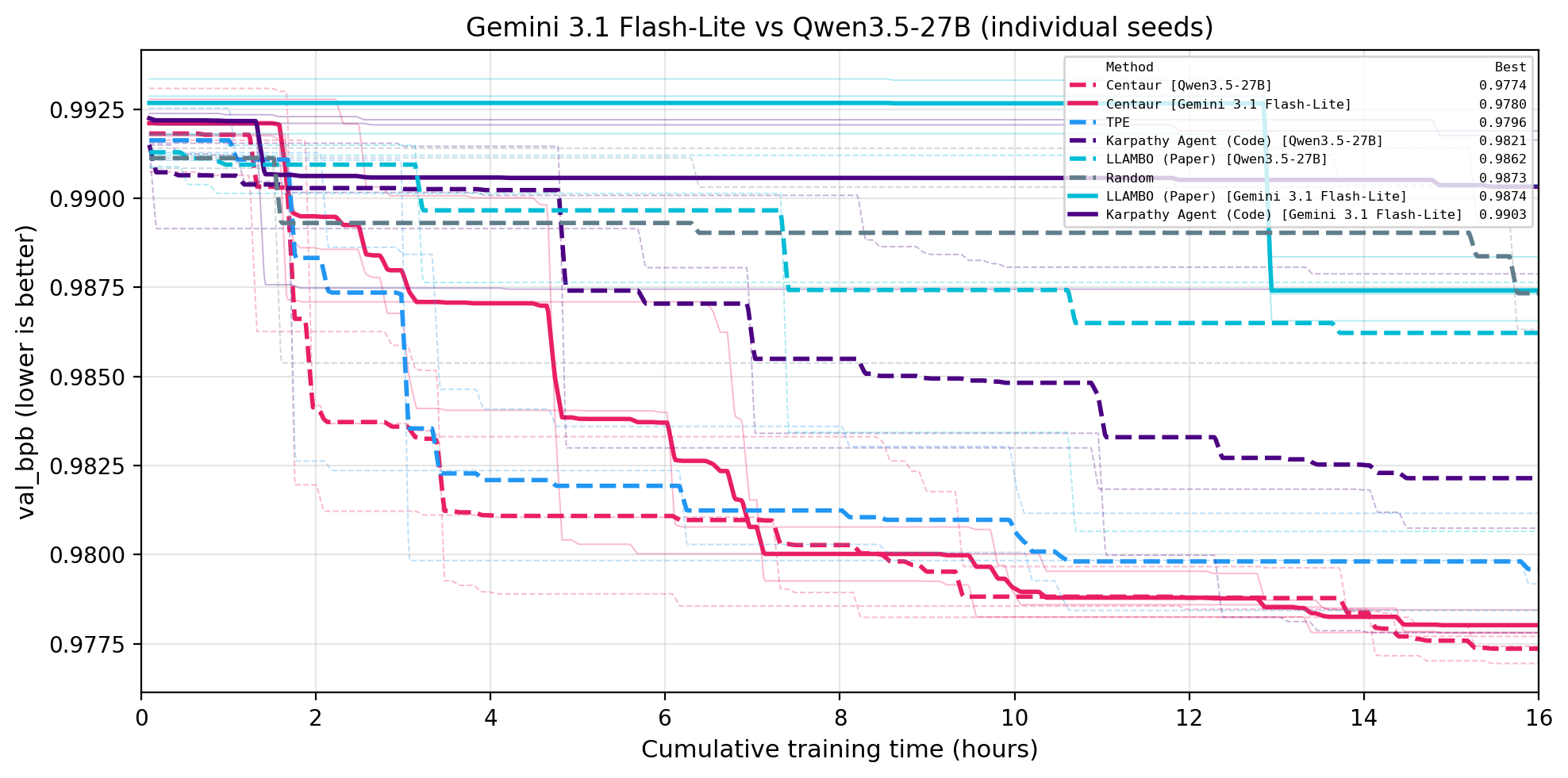}
    \caption{Gemini 3.1 Flash-Lite vs Qwen3.5-27B as LLM optimizer (cumulative training time). Solid: Gemini 3.1, dashed: Qwen3.5-27B. Same color per method.}
    \label{fig:gemini31}
\end{figure}

\Cref{fig:gemini,fig:gemini31} compare the Flash variants with Qwen3.5-27B, but neither outperforms it. Notably, Karpathy Agent (Code) with Gemini 3.1 Flash-Lite suffers from extremely high failure rates (87--94\%), as the model frequently generates code modifications that cause the training script to crash (93\% of failures are runtime errors, not OOM). In the fixed search space, however, Gemini 3.1 Flash-Lite performs comparably to Qwen3.5-0.8B. \Cref{fig:frontier_main} shows that Gemini 3.1 Pro Preview is competitive with Qwen3.5-27B for both unconstrained code editing (Karpathy Agent Code: 0.9826${\pm}$0.0004 vs 0.9814${\pm}$0.0046, 3\% OOM) and \centaur in the fixed search space (0.9767${\pm}$0.0013 vs 0.9763${\pm}$0.0005). Gemini Pro achieves lower variance than Qwen for code editing but does not improve the mean, confirming that scaling the LLM optimizer does not close the gap to classical methods.

\subsection{Centaur LLM Ratio Ablation}
\label{sec:ratio_ablation}

We ablate the fraction of trials delegated to the LLM in \centaur. \Cref{tab:ratio_ablation} reports best val\_bpb for each ratio. The default ratio of $r{=}0.3$ balances classical and LLM contributions; higher ratios degrade performance, particularly for the 27B model where $r{=}0.8$ performs worse than CMA-ES alone. This confirms that CMA-ES needs to retain majority control of the optimization trajectory, with the LLM contributing occasional informed suggestions rather than dominating the search.

\begin{table}[ht]
\centering
\caption{Centaur LLM ratio ablation. Best val\_bpb (mean${\pm}$std). 2 seeds per ratio, 3 seeds for default $r{=}0.3$. Bold = best per column.}
\label{tab:ratio_ablation}
\small
\begin{tabular}{@{}ccc@{}}
\toprule
LLM Ratio $r$ & \centaur [0.8B] & \centaur [27B] \\
\midrule
0.1 & \textbf{0.9744${\pm}$0.0003} & 0.9758${\pm}$0.0005 \\
0.2 & 0.9748${\pm}$0.0013 & 0.9748${\pm}$0.0005 \\
0.3 (default) & 0.9766${\pm}$0.0008 & 0.9763${\pm}$0.0005 \\
0.5 & 0.9768${\pm}$0.0019 & \textbf{0.9746${\pm}$0.0000} \\
0.8 & 0.9849${\pm}$0.0060 & 0.9902${\pm}$0.0000 \\
\bottomrule
\end{tabular}
\end{table}

\begin{figure}[ht!]\vspace{0.5em}
    \centering
    \includegraphics[width=0.75\textwidth]{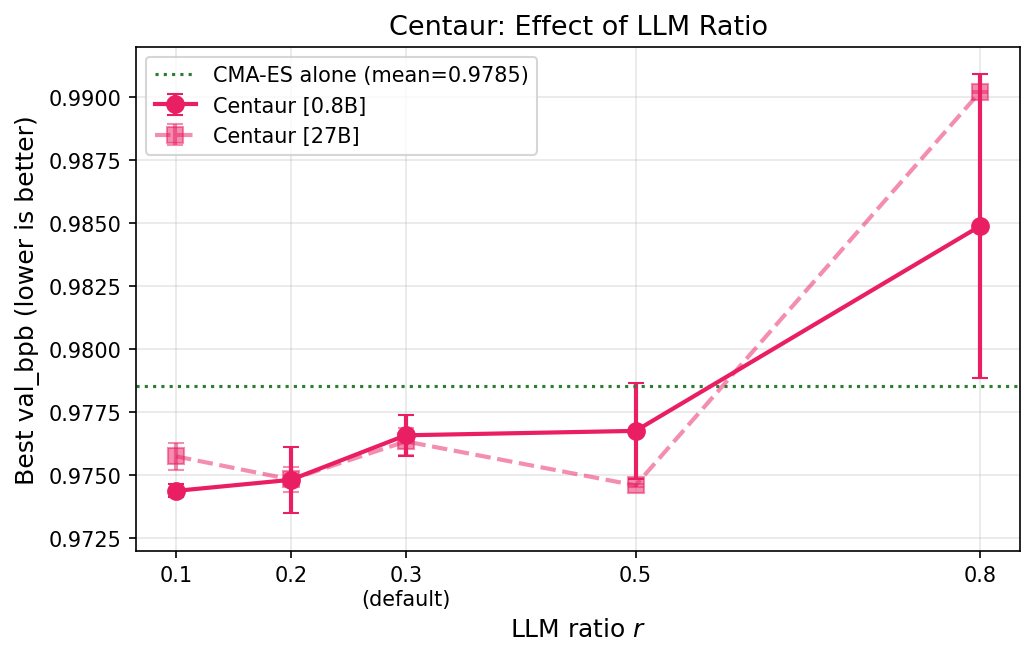}
    \caption{Effect of LLM ratio on \centaur performance (2 seeds, 3 for default). Higher ratios give the LLM more trials. CMA-ES alone (dotted) shown as reference. Too much LLM control degrades performance, especially for the 27B model at $r{=}0.8$.}
    \label{fig:ratio_ablation}
\end{figure}

\begin{figure}[ht!]\vspace{0.5em}
    \centering
    \includegraphics[width=\textwidth]{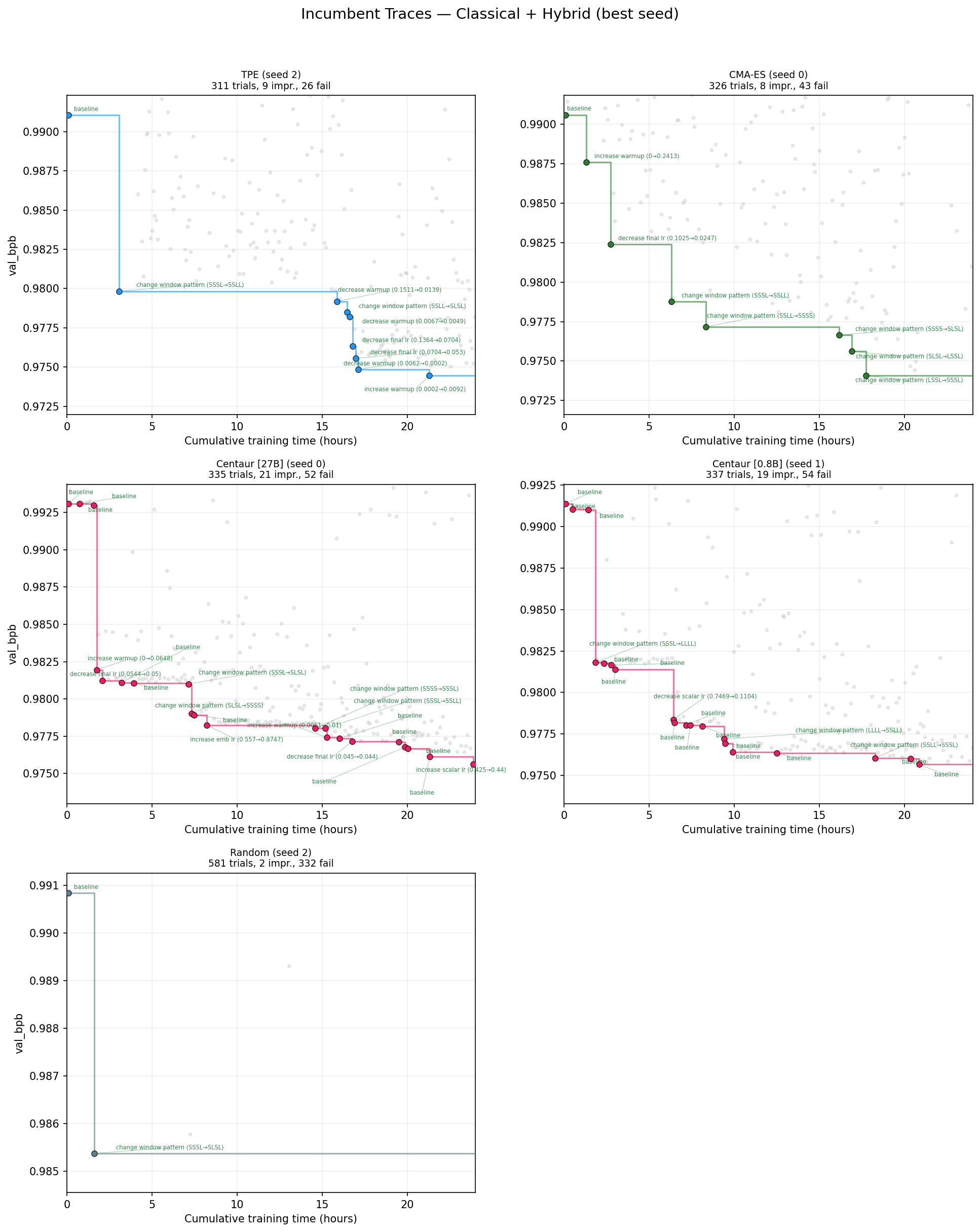}
    \caption{Incumbent traces for classical and hybrid methods (wall-time, best seed). Grey dots: all trials. Colored dots: new incumbents. Staircase: best-so-far.}
    \label{fig:incumbents_classical}
\end{figure}

\subsection{Incumbent Traces}

\begin{figure}[ht!]\vspace{0.5em}
    \centering
    \includegraphics[width=\textwidth]{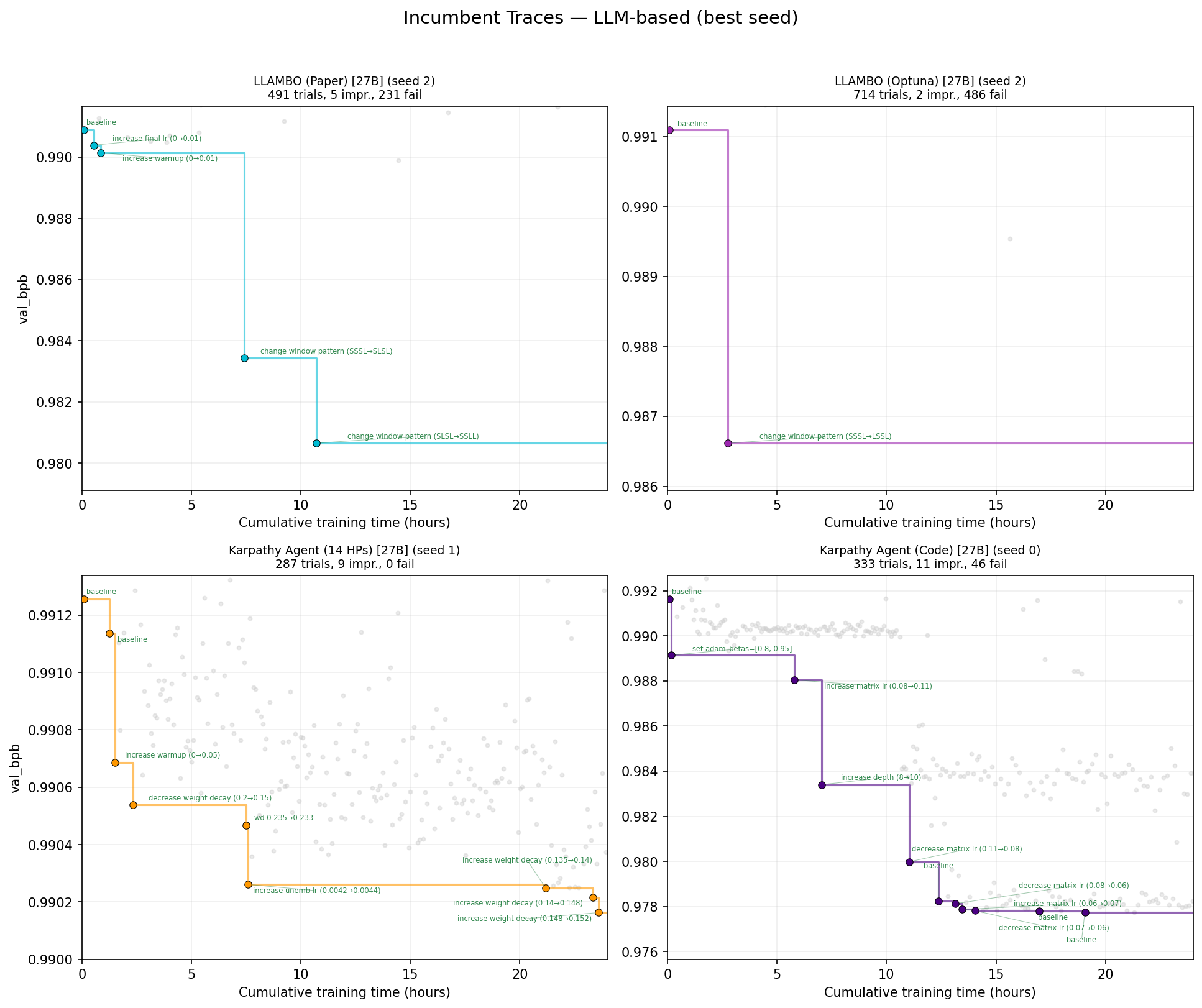}
    \caption{Incumbent traces for LLM-based methods (wall-time, best seed).}
    \label{fig:incumbents_llm}
\end{figure}

Beyond aggregate convergence, we visualize per-method incumbent traces to reveal when and how each optimizer finds improvements. \Cref{fig:incumbents_classical,fig:incumbents_llm} show incumbent traces against cumulative training time (hours) for all methods. Grey dots are all trials, colored dots are new incumbents, and the staircase line is the best-so-far trajectory. Each panel shows the best seed for that method.

\subsection{Qualitative Agent Behavior}
\label{sec:centaur_analysis}

The convergence plots show that \centaur improves over CMA-ES alone; we now examine how the LLM uses the shared optimizer state in practice. In \centaur seed~0, trial~136 produced a new incumbent (val\_bpb${}=0.9837$): 
\begin{itemize}[nosep,leftmargin=*]
    \item \textbf{CMA-ES suggested:} \texttt{WINDOW\_PATTERN=LLLL}, \texttt{DEVICE\_BATCH\_SIZE=61}, \texttt{TOTAL\_BATCH\_SIZE=133143}, \texttt{SCALAR\_LR=0.208}
    \item \textbf{LLM overrode to:} \texttt{WINDOW\_PATTERN=SSSS}, \texttt{DEVICE\_BATCH\_SIZE=64}, \texttt{TOTAL\_BATCH\_SIZE=131072}, \texttt{SCALAR\_LR=0.3}
\end{itemize}

Three overrides are notable: (1)~\textbf{Attention pattern} (LLLL$\to$SSSS): CMA-ES has no domain knowledge about attention patterns. The LLM chose all-short attention, which is memory-efficient at the given depth (DEPTH${}=10$). This is transformer-specific knowledge that CMA-ES cannot learn from scalar loss values. (2)~\textbf{Hardware-friendly rounding}: the LLM chose power-of-2 batch sizes (64, 131072) instead of CMA-ES's arbitrary values (61, 133143), aligning with GPU memory and tensor core constraints. (3)~\textbf{Learning rate}: the LLM boosted SCALAR\_LR from 0.208 to 0.3, closer to the regime where good configs cluster.

Overall in seed~0, 6 of 24 incumbent improvements came from LLM trials (25\%), while LLM trials constituted 88 out of 275 total trials (32\%, close to the 30\% target ratio).

\subsection{Diversity Metrics}
\label{sec:diversity_metrics}

We now define the diversity metrics reported in \Cref{tab:diversity}. We compute all metrics on the 13 continuous HPs, excluding the categorical \texttt{WINDOW\_PATTERN}. We normalize each HP to $[0,1]$ within its bounds and use only successful (non-OOM) trials.

\begin{itemize}[nosep,leftmargin=*]
\item \textbf{Spread}: mean standard deviation per HP across all trials. Higher values indicate more diverse sampling across each dimension.
\item \textbf{Pairwise}: mean L2 distance between all pairs of configurations. Higher values indicate configs are more different from each other.
\item \textbf{Step}: mean L2 distance between consecutive trials. Higher values indicate larger jumps between suggestions.
\item \textbf{Cells}: we discretized each HP into 5 equal-width bins and counted the number of unique 13-dimensional bin vectors across all trials. The theoretical maximum is $5^{13} \approx 1.2 \times 10^9$; values in \Cref{tab:diversity} range from 29 to 805, indicating that all methods cover a small fraction of the search space.
\end{itemize}

\subsection{0.8B LLM Variant Results}

\Cref{tab:diversity_08b} reports diversity metrics for the remaining 0.8B LLM variants not shown in the main \Cref{tab:diversity}, i.e., the fixed-search-space pure LLM methods.

\begin{table}[ht]
\centering
\caption{Diversity analysis for 0.8B LLM variants of fixed-search-space pure LLM methods (3 seeds each). Same columns as \Cref{tab:diversity}.}
\label{tab:diversity_08b}
\scriptsize
\begin{tabular}{@{}llclcc@{}}
\toprule
Method & Trials & Best val\_bpb & OOM\% & Spread & Step \\
\midrule
Karpathy Agent (14 HPs) & 338${\pm}$3 & \textbf{0.9908${\pm}$0.0002} & 16\% & 0.037 & 0.062 \\
LLAMBO (Optuna) & 668${\pm}$29 & 0.9873${\pm}$0.0003 & 64\% & 0.297 & 1.311 \\
LLAMBO (Paper) & 605${\pm}$6 & 0.9894${\pm}$0.0007 & 59\% & 0.272 & 1.330 \\
\bottomrule
\end{tabular}
\end{table}

\subsection{LLAMBO (Optuna) vs LLAMBO (Paper): Detailed Comparison}
\label{sec:llambo_comparison}

We include two LLAMBO variants because the OptunaHub port differs from the original paper in ways that substantially affect performance. \Cref{tab:llambo_diff} summarizes the key implementation differences.

\begin{table}[ht]
\centering
\caption{Key differences between LLAMBO (Optuna) and LLAMBO (Paper).}
\label{tab:llambo_diff}
\small
\begin{tabular}{@{}lll@{}}
\toprule
Aspect & Original Paper & OptunaHub Port \\
\midrule
Surrogate labels & Continuous metric values & Binary 0/1 (top 20\% threshold) \\
Categorical HPs & All HPs in LLM prompts & Delegated to random sampling \\
Failed trials & Visible to surrogate & Hidden (\texttt{TrialState.FAIL}) \\
\bottomrule
\end{tabular}
\end{table}

\subsection{Full Convergence with All Methods}

\begin{figure}[ht!]\vspace{0.5em}
    \centering
    \includegraphics[width=\textwidth]{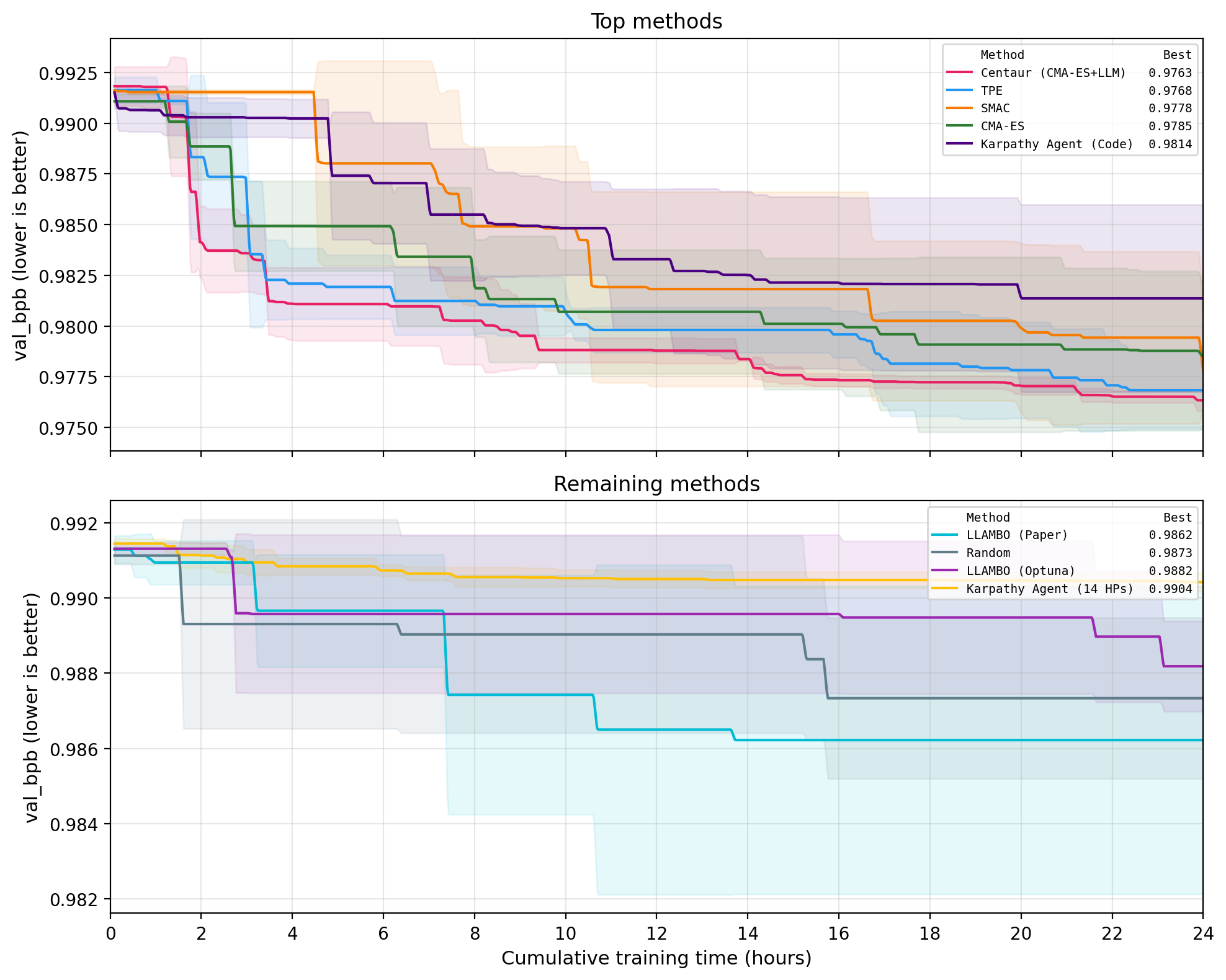}
    \caption{All 9 methods with $\pm$std bands, split into top 5 (Centaur, TPE, SMAC, CMA-ES, Karpathy Agent Code) and remaining 4 (Karpathy Agent 14 HPs, LLAMBO Paper, LLAMBO Optuna, Random).}
    \label{fig:twopanel}
\end{figure}

\Cref{fig:convergence} shows $\pm$std bands for the top 5 methods only. \Cref{fig:twopanel} shows all 9 methods with bands, split into competitive methods (top) and remaining methods (bottom) for readability.

\clearpage
\subsection{LLM Prompts and Problem Context}
\label{sec:prompts}

We report the prompt templates used by all five LLM-based methods in our benchmark. Each prompt includes the optimization goal, the model class and training stack, the dataset, the hardware constraints and OOM warning, the search space with bounds, and the trial history. Braces of the form \texttt{\{name\}} are Python format placeholders filled in at runtime. These templates are also available in our public code repository\footnote{\url{https://github.com/ferreirafabio/autoresearch-automl}} in the respective backend files.

\paragraph{Karpathy Agent (Code).} Source: \texttt{autoresearch\_automl/backends/karpathy\_agent\_backend.py}. The LLM receives the current best \texttt{train.py} source code together with problem context and returns a full replacement for \texttt{train.py}.
\begin{lstlisting}[basicstyle=\ttfamily\scriptsize,breaklines=true,frame=single]
You are an autonomous ML researcher. Your goal: minimize val_bpb (validation bits-per-byte) for a GPT-2 scale transformer trained on climbmix-400b-shuffle.

Current train.py (the version that achieved the best result so far):
```python
{current_source}
```

Previous experiments:
{history}

Current best val_bpb: {best_val_bpb}

Rules:
- You can change ANYTHING in train.py: architecture, optimizer, hyperparameters, training loop, model size.
- Do NOT modify imports from prepare (prepare.py is read-only).
- The script runs for exactly 5 minutes on a single GPU with {available_vram} VRAM available.
- Configs that use too much memory will crash (OOM) - be mindful of DEPTH, HEAD_DIM, DEVICE_BATCH_SIZE.
- Simplicity wins: a small improvement from deleting code is better than a large improvement from adding complexity.
- Think about what previous experiments tell you - learn from both successes and failures.

Respond in EXACTLY this format (no extra text before or after):

DESCRIPTION: <one line explaining what you changed and why>
```python
<the complete modified train.py - must be valid, runnable Python>
```
\end{lstlisting}

\paragraph{Karpathy Agent (14 HPs).} Source: \texttt{autoresearch\_automl/backends/karpathy\_agent\_hps\_backend.py}. The LLM is constrained to the fixed 14-HP search space and must return a JSON configuration.
\begin{lstlisting}[basicstyle=\ttfamily\scriptsize,breaklines=true,frame=single]
You are optimizing hyperparameters for a GPT-2 scale transformer trained on climbmix-400b-shuffle.
The goal is to minimize val_bpb (validation bits-per-byte).
Training runs on a single H200 GPU (141GB VRAM, {available_vram} available after vLLM) - configs with very high DEPTH, large HEAD_DIM,
or large DEVICE_BATCH_SIZE may cause OOM (out of memory) crashes.

Current search space with bounds:
{space_description}

Previous evaluations (config -> result):
{history}

{incumbent_info}

Suggest the next configuration to try. Learn from both successful runs AND failures (OOM crashes
mean the model was too large for the GPU). Respond with a JSON object mapping HP names to values.
Only include HPs from the search space. Be creative but principled - consider what you know
about transformer training dynamics.

Respond ONLY with a valid JSON object, no explanation.
\end{lstlisting}

\paragraph{Centaur (CMA-ES + LLM).} Source: \texttt{autoresearch\_automl/backends/centaur\_backend.py}. On the 30\% of trials where the LLM overrides CMA-ES, it receives CMA-ES's internal state (mean, sigma, covariance summary, top-5 configs, last 20 trials) alongside the problem context.
\begin{lstlisting}[basicstyle=\ttfamily\scriptsize,breaklines=true,frame=single]
You are optimizing hyperparameters for a GPT-2 scale transformer trained on climbmix-400b-shuffle.
Goal: minimize val_bpb. GPU: H200 with {available_vram} available.

Search space:
{space_description}

{cma_analysis}

Previous evaluations (last 20):
{history}

{incumbent_info}

Use CMA-ES's analysis as guidance:
- If CMA-ES is converging (low sigma), exploit near its mean
- If sigma is large, CMA-ES is still exploring - help by using your transformer knowledge
- You can diverge from CMA-ES if you see patterns it might miss (e.g., OOM-prone regions, known good ratios)

Respond ONLY with a valid JSON object mapping HP names to values.
\end{lstlisting}

\paragraph{LLAMBO (Paper) and LLAMBO (Optuna).} Sources: \texttt{autoresearch\_automl/backends/llambo\_original\_backend.py} and \texttt{llambo\_backend.py}. Both variants build their prompts via the LLAMBO surrogate / acquisition templates from \citet{liu2024llambo}, with a custom task description prefixed that we reproduce below. LLAMBO (Optuna) is the \citep{ozaki2025optunahub} port; LLAMBO (Paper) is our faithful reimplementation of the original (details in \Cref{sec:llambo_comparison}).
\begin{lstlisting}[basicstyle=\ttfamily\scriptsize,breaklines=true,frame=single]
Optimizing a GPT-2 scale transformer for language modeling on climbmix-400b-shuffle.
The model uses a Muon+AdamW optimizer with weight decay and learning rate scheduling.
Architecture hyperparameters control model depth, width (via aspect ratio and head dim), and attention patterns. Optimization hyperparameters control learning rates for different parameter groups (embedding, unembedding, matrix, scalar), weight decay, and batch sizes.
Goal: minimize val_bpb (validation bits-per-byte).
Training runs on a single H200 GPU (141GB VRAM, {available_vram} available after vLLM) with a fixed time budget of {budget}s.
VRAM is a soft constraint - large models with high DEPTH and large DEVICE_BATCH_SIZE can OOM.
WINDOW_PATTERN is encoded as ordinal: 0=SSSL, 1=SSLL, 2=SLSL, 3=LLLL, 4=SSSS, 5=LSSL (controls sliding window vs full attention pattern per layer).
\end{lstlisting}

The LLAMBO surrogate and acquisition functions then append the hyperparameter constraints, few-shot examples drawn from prior trials, and a query for the next configuration. LLAMBO (Paper) uses continuous metric values as surrogate labels and keeps failed trials visible; LLAMBO (Optuna) uses binary top-20\% labels and hides failed trials (see \Cref{tab:llambo_diff}).

\end{document}